\documentclass{article}
\usepackage{spconf,amsmath,epsfig}
\usepackage{hyperref}
\usepackage{booktabs}
\let\OLDthebibliography\thebibliography
\renewcommand\thebibliography[1]{
  \OLDthebibliography{#1}
  \setlength{\parskip}{0pt}
  \setlength{\itemsep}{0pt plus 0.3ex}
}

\begin{document}\sloppy

\def\x{{\mathbf x}}
\def\L{{\cal L}}

\title{SSHR: Leveraging Self-supervised Hierarchical Representations for Multilingual Automatic Speech Recognition}
%
\name{Hongfei Xue$^{1}$, Qijie Shao$^{1}$, Kaixun Huang$^{1}$, Peikun Chen$^{1}$,  Jie Liu$^{2}$, Lei Xie$^{1*}$\thanks{*Corresponding author.}}
\address{$^1$Audio, Speech and Language Processing Group (ASLP@NPU), \\ School of Software, Northwestern Polytechnical University, Xian, China \\
  $^2$Huawei Cloud, China}

\maketitle

\begin{abstract}
Multilingual automatic speech recognition (ASR) systems have garnered attention for their potential to extend language coverage globally. 
While self-supervised learning (SSL) models, like MMS, have demonstrated their effectiveness in multilingual ASR, it is worth noting that various layers' representations potentially contain distinct information that has not been fully leveraged.
In this study, we propose a novel method that leverages self-supervised hierarchical representations (SSHR) to fine-tune the MMS model. 
We first analyze the different layers of MMS and show that the middle layers capture language-related information, and the high layers encode content-related information, which gradually decreases in the final layers.
Then, we extract a language-related frame from correlated middle layers and guide specific language extraction through self-attention mechanisms.
Additionally, we steer the model toward acquiring more content-related information in the final layers using our proposed Cross-CTC.
We evaluate SSHR on two multilingual datasets, Common Voice and ML-SUPERB, and the experimental results demonstrate that our method achieves state-of-the-art performance. 

\end{abstract}
\begin{keywords}
Multilingual ASR, self-supervised learning, representation analysis, low-resource ASR.
\end{keywords}
\section{Introduction}
\label{sec:intro}
With more than 7,000 languages around the globe, most languages still lack adequate support from speech technologies~\cite{mms}. In recent years, both the academic and industrial communities have displayed considerable interest in multilingual automatic speech recognition (ASR)~\cite{18Multi, 20MMASR, 21scaling, 21XLSR, 22JUST, 22Whisper} to expand language coverage. These studies can be broadly categorized into two types: one involves supervised or semi-supervised learning of end-to-end ASR systems using multilingual data~\cite{18Multi, 19Largescale, 20MMASR, 21scaling, 22Device, 22Whisper}, while the other utilizes self-supervised learning (SSL) techniques to learn meaningful multilingual generalized representations from vast amounts of unlabeled data~\cite{21XLSR, 21UniSpeech, 22JUST, 22XLS-R, mms, xue2023tranusr, 23USM}. Specifically, the latter utilizes SSL to create generalized representations, which often exhibit superior performance when lacking enough labelled data in low-resource languages.

While fine-tuning SSL for downstream tasks is a simple and effective approach, research has proven that limited relevant information is available at the final layers of the SSL model~\cite{21layerwise}. Research in SSL representations has unveiled a notable correlation between the middle layers and language-related information~\cite{22LID}. Additionally, the middle and high layers tend to encapsulate more content-related information~\cite{21layerwise}. However, this content-related information diminishes as we progress through the model's final layers. Although the effectiveness of current SSL models like Massively Multilingual Speech (MMS)~\cite{mms} have been proven in multilingual ASR, the extent to which the layers of the SSL model contribute to the specific task still needs to be explored. Consequently, the optimal utilization of SSL's hierarchical representations to enhance the fine-tuning performance of downstream multilingual ASR tasks presents an unresolved challenge.  In our study, hierarchical representations refer to the multi-layered representations within SSL models that are employed to enhance outcomes specific to a task.

The complexity of constructing a multilingual ASR system arises from the need to accommodate significant variations in acoustic, linguistic, and semantic across diverse languages~\cite{23HierCTC}. Consequently, the key to achieving a successful multilingual ASR system is ensuring the model can accurately recognize and transcribe specific languages. This can be achieved by exploring language-related information from SSL's middle layers. With language identification (LID) in place, the subsequent challenge is utilizing the content-related information to perform downstream ASR tasks more accurately. To achieve satisfactory results in fine-tuning, the final layers of the SSL must contain substantial content-related information.

This paper proposes \textbf{s}elf-\textbf{s}upervised \textbf{h}ierarchical \textbf{r}epresentations (SSHR) to improve multilingual ASR performance. Specifically, our approach encompasses three key refinements during the fine-tuning process of MMS~\cite{mms}. First, we extract a LID-related frame from the middle layers and concat it into encoder frames to guide specific language content extraction in the subsequent layers. 
Second, considering diminishing content-related information in the final layers, we can fine-tune MMS by introducing Connectionist Temporal Classification (CTC)~\cite{20ctc} in the higher content-related information layer.
Finally, to further enhance performance, we propose Cross-CTC to acquire more content-related information in the final layers.
We evaluate the performance of the SSHR on two multilingual datasets, ML-SUPERB~\cite{23MLSUPERB} and Common Voice~\cite{19commonvoice}. Our contributions can be summarized as follows: (1) We analyze the layer-wise representations of MMS and discover that the middle layers contain more language-related information. The middle and high layers contain more content-related information, but the final layers will be lost.
(2) We propose the SSHR, which leverages the SSL hierarchical representations to improve the performance of downstream multilingual ASR. (3) Compared with the baseline, our method achieves relative 9.4\% and 12.6\% performance improvements on both datasets, reaching the state-of-the-art (SOTA) performance.

\section{Related Work}
\label{sec:related_work}

\textbf{Leveraging Language-related Information. }
In the context of multilingual ASR, various approaches have been explored to predict LID in the middle of the model. Some methods attempt to predict languages in a streaming manner within end-to-end ASR and integrate them into the decoding process~\cite{22jointLID, 22Device}. Moreover, HierLID~\cite{23HierCTC} introduces encoder and decoder layers following SSL features, training lower encoders specifically for language classification and incorporating predictions into subsequent layers. Compared to HierLID, our SSHR not only extracts language-related information from the middle layers to enhance the model's performance in the following tasks but also performs LID prediction at the final layer without introducing errors. SSHR applies directly to pre-trained models, diverging from HierLID's method of using a frozen pre-trained model supplemented by encoder-decoder for intermediate representations. Crucially, SSHR avoids mid-process LID prediction, thereby eliminating potential prediction errors.

\textbf{Improving Content-related Information. }
To address the issue of missing content-related information in the final layers, one method proposes a strategy of random initialization followed by fine-tuning in the final three layers to achieve better results~\cite{21layerwise}. In end-to-end ASR, the SC-CTC~\cite{21scctc} method incorporates the posterior probabilities of Inter-CTC~\cite{21interctc} into the input of the rear layer to provide textual information. Furthermore, GIC~\cite{22GIC} employs an additional gating network to more deeply fuse textual and speech data, aiming to counteract CTC's inherent independence assumptions. SSHR extends these approaches by incorporating them into the pre-trained model, thereby augmenting content information in final layers. Moreover, our novel cross-CTC approach explicitly and effectively guides the learning of content information in final layers, distinguishing it from existing techniques.


\section{The Proposed Method}
\label{sec:proposed_method}
\begin{figure}[h]

\centering
\includegraphics[width=0.9\linewidth]{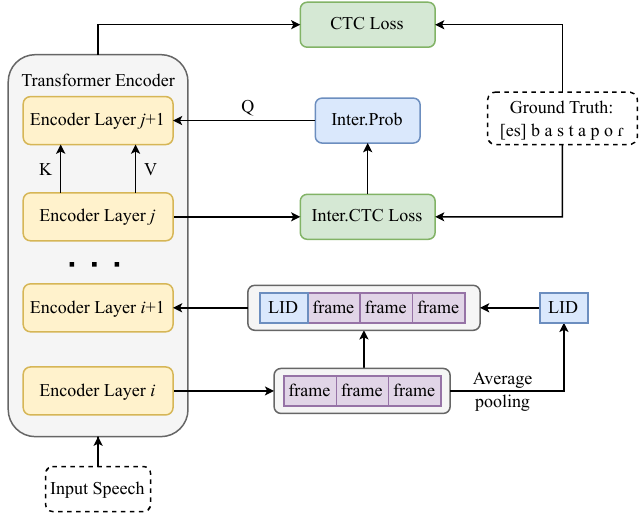}
\caption{
The overall framework of the proposed SSHR. The Inter.CTC Loss is the CTC loss of the $j$-th layer, while the Inter.Prob is the posterior probabilities of the Inter.CTC Loss.}
\label{fig:model}
\vspace{-20pt}
\end{figure}

We aim to harness the hierarchical representations in multilingual SSL to improve the downstream multilingual ASR tasks. The overarching model architecture is illustrated in Fig.\ref{fig:model}. Within this section, we intricately unpack the specifics of each component comprising the model.

\subsection{Layer-wise Analysis of MMS}
MMS~\cite{mms} follows Wav2vec 2.0's~\cite{20wav2vec} self-supervised pre-training process, which only requires unlabeled speech and can be adapted for ASR with a small amount of labeled data. Our model is based on MMS, built pre-trained on 491k hours of speech data covering 1,406 languages. 
We analyze the layer-wise representations of MMS, focusing on both language-related and content-related aspects. Regarding language-related information, we employ logistic regression training across each layer of MMS representations, followed by an assessment of language prediction accuracy using a designated test set. Regarding content-related information, we employ $K$-means clustering on the representations to compute the mutual information (MI) between hierarchical representations and the corresponding phoneme labels~\cite{21layerwise}.

\subsection{Extracting Language-related Information}
As illustrated in Fig.\ref{fig:model}, we employ an average-pooling operation within SSHR on the output of the $i$-th layer, which contains rich language-related information. The average-pooling operation reduces it to one dimension, representing a LID-related frame. 
We splice this one-dimensional information with speech frames and then apply self-attention in the following layers to continuously reinforce the language-related information in the LID-related frame and use the LID-related frame to guide learning other speech frames.
Ultimately, we utilize CTC loss to control this LID dimension.
For models with the LID-related frame, the ground truth is ``$[LID], phoneme, ..., phoneme$"  labels, or just phoneme labels if no LID-related frame is used. The CTC loss can be expressed as:
\begin{align}
L_{\text{ctc}}^{j} = \sum_{t=1}^{T+1} \log P_{\text{CTC}}(y_t | \mathbf{X}^{j}),
\end{align}
where $y_1$ denotes the LID label, $j$ represents the $j$-th layer.
It is important to note that our method does not involve a separate Cross-Entropy loss for the LID frame. Rather, the CTC loss automatically aligns the LID label with the first frame for LID identification.


\subsection{Refining SSL Parameters}
\label{sc:3,3}
We delve into adjusting SSL parameters to enhance the fine-tuning process for improved ASR performance. A primary concern lies in the potential impact on fine-tuning quality, attributed to the reduced information available concerning the content of the post-$n$ layers~\cite{21layerwise}. To address this challenge, we explore two distinct approaches.

1) Replace final $n$ layers. Recognizing the deficiency in information within the post-n layer, we opt for a solution that leverages intermediary layers with richer content-related information. These intermediate layers serves as the source for initializing the post-$n$ layer. More specifically, if $n$ equals 3, we redefine the model structure as [1, 2, 3, ... 19, 20, 21, 19', 20', 21'].

2) Delete final $n$ layers. We propose a departure from optimizing the latter $n$ layers, which inherently possess less content-related information. A more promising avenue is performing supervised CTC fine-tuning on the penultimate $n$+1 layers and abandoning the last $n$ layers, characterized by more substantial content-related information. This innovative approach redefines the model structure as [1, 2, 3, ... $24-n$].

\subsection{Enhancing Content-related Information}
\label{sc:3,4}
In Section \ref{sc:3,3}, we employ the delete final $n$ layers approach to remove the last few layers with insufficient content-related information. Nonetheless, to further enhance performance, we propose a Cross-CTC method designed to augment the information in these final layers.

We incorporate CTC loss within the intermediate layers to derive posterior probabilities. As illustrated in Fig.\ref{fig:model}, instead of directly appending the posterior probabilities to the outputs of layer $j$~\cite{22GIC}, we employ cross-attention with the output of layer $j$ serving as `$K$' and `$V$', and the linear transformation of the posterior probabilities as `$Q$'. The posterior probabilities of content-related information from layer $j$ are integrated into the output of layer $j$+1 as `$Q$'. Consequently, the output of layer $j$+1 is guided by content-related information, naturally reinforcing the content-related information within the subsequent $n$ layers. The specific calculation process proceeds as follows:
\begin{align}
\mathbf{X}^{j+1} &= \text{Encoder Layer}(\mathbf{Q}^{j}, \mathbf{K}^{j}, \mathbf{V}^{j}) ,\\
\mathbf{Q}^{j} &= \text{Linear}(\text{Linear}(\mathbf{P}^{j})) ,\\
\mathbf{K}^{j} &= \text{Linear}(\mathbf{X}^{j}), \quad \mathbf{V}^{j} = \text{Linear}(\mathbf{X}^{j}), 
\end{align}
where $P^j$ is posterior probabilities from layer $j$, $X^j$ is output representations from layer $j$.

Similar to methods like GIC~\cite{22GIC}, our Cross-CTC utilizes multiple intermediate layers to calculate the CTC losses. Nevertheless, what distinguishes Cross-CTC is our method of choosing the cross-attention implementation layer based on the specific representations of each layer. Our method chooses those that exhibit higher content-related information instead of applying inter-CTC uniformly to multiple layers. This allows for model-targeted optimization. The final loss is as follows:
\begin{align}
L_{\text{all}} = (1 - w) \cdot L_{\text{ctc}} + w \cdot \frac{1}{k} \sum_{j=j_1}^{j_k} L_{\text{ctc}}^j
\end{align}
where $[j_1,...,j_k]$ are higher content-related information layers.

\section{Experiments}
\label{sec:experiments}
\subsection{Datasets}
In this study, we have chosen two datasets, namely Common Voice\footnote{https://commonvoice.mozilla.org/en/datasets. We utilized the December 2020 release for model training.}~\cite{19commonvoice} and ML-SUPERB~\cite{23MLSUPERB} to comprehensively evaluate the generalization capacity of our proposed method across diverse scenarios. The Common Voice dataset incorporates eight languages, in contrast to the 143 languages included in ML-SUPERB, validating the robustness and applicability of our model.

Within the Common Voice dataset, we make a selection of languages following prior works\cite{21UniSpeech, xue2023tranusr}, encompassing Spanish, French, Italian, Kyrgyz, Dutch, Russian, Swedish, and Tatar. Notably, each language contributes training, validation, and test sets, each consisting of 1, 0.5, and 1 hour of speech data, respectively. We leverage the phonemizer tool\footnote{https://github.com/bootphon/phonemizer} for converting text to phonemes to facilitate comparisons with prior research, and the phoneme set is characterized by a size of 139. In addition, ML-SUPERB encapsulates 143 languages, spanning the spectrum from high-resource to endangered languages. For our experimental framework, we harness the provided 1-hour set from ML-SUPERB\footnote{https://multilingual.superbbenchmark.org/}. The grapheme set is characterized by a size of 5000.

\subsection{Implementation Details}
\textbf{Model Architecture.} Our experimental framework is conducted on the fairseq ~\cite{19fairseq}, utilizing input data in the 16K Hz rate, and is executed across 4 GPUs (4090, 24G). Anchored in the MMS 300M architecture, our model comprises 24 transformer encoder layers. Moreover, we set the weight $w$ to 0.5. It is important to emphasize that incorporating the LID-related frame does not introduce supplementary parameter overhead. Simultaneously, a linear CTC component, equipped with parameter-sharing capabilities, is integrated into the intermediate layer via Cross-CTC. In comparison to the original configuration, this augmentation results in an increase of approximately 5M parameters in the model.

\textbf{Common Voice~\cite{19commonvoice}. }For the Common Voice dataset, we draw insights from the previous benchmark results of XLSR-10 and XLSR-53~\cite{21XLSR}. Leveraging the MMS 300M model, we perform direct fine-tuning over the joint 8-hour dataset in 8 languages as our baseline. We employ Adam optimization~\cite{14adam} during the fine-tuning of MMS, using a peak learning rate of 5e-5 and a gradient accumulation factor of 4 for 20k steps. We use a batch of around 1M samples on each GPU.

\textbf{ML-SUPERB~\cite{23MLSUPERB}. }We utilize the XLSR-128~\cite{22XLS-R} model as a benchmark reference on the ML-SUPERB dataset, with the encoder parameters kept frozen during fine-tuning~\cite{23MLSUPERB}. Leveraging the MMS 300M model, we directly fine-tune using labeled data encompassing 143 languages as the baseline, and the encoder is not frozen. We fine-tune MMS for 80k steps with a learning rate 3e-5 and a gradient accumulation factor of 4. We use a batch of around 1.6M samples on each GPU.
\vspace{-2pt}

\subsection{Experimental Results}
The results of the SSHR method on the two datasets are presented in Table~\ref{tab:SSHR}. We evaluate the Common Voice dataset using the phoneme error rate (PER). In contrast, we assess both the word error rate (WER) and the character error rate (CER) for the ML-SUPERB dataset. B0 is the baseline model derived from our direct fine-tuning of the MMS. Meanwhile, C1 to C4 denote the various experiments we conducted on the SSHR model to investigate their impact. In C1, we involve the extraction of the LID-related frame utilizing the representation from layer 8. In C2, we apply the delete $n$ final layers technique, wherein $n$ corresponds to layer 3. In C3, we utilize the Cross-CTC in layers 18, 20, and 22 to enrich the content-related information of subsequent layers, respectively. Finally, C4 embodies the comprehensive SSHR, with relative improvements of 9.4\%, 12.6\%, and 9.5\% compared to the baseline model on both datasets, respectively. Our model achieves SOTA results, demonstrating an improvement over the previous XLSR works. These results underline the effectiveness of our proposed SSHR method in discernibly improving the performance of multilingual ASR tasks through the utilization of hierarchical layer representations.
\vspace{-5pt}
\begin{table}[h]
\centering
\caption{SSHR on Common Voice and ML-SUPERB. `\textbf{*}' means eight languages are not joint fine-tuned. CV-PER represents PER on Common Voice, ML-CER represents CER on ML-SUPERB, and ML-WER represents WER.}
\label{tab:SSHR}
\resizebox{1.0\linewidth}{!}{
\begin{tabular}{lllll}
\toprule
ID   & Model            & CV-PER & ML-CER & ML-WER \\ \midrule
P1    & XLSR-10~\cite{21XLSR}  & 11.2    &  -     &  -      \\
P2    & XLSR-53~\cite{21XLSR} & 6.36\textbf{*}    &  26.9     &   -     \\
P3    & XLSR-128~\cite{23MLSUPERB}        &  -     & 22.0   &  -      \\ \midrule
B0   & Baseline         & 6.70   & 16.08  & 50.14  \\
C1 & \hspace{1em}+ LID-related frame  & 6.39   & 15.08  & 47.81  \\
C2 & \hspace{1em}+ Delete final layers     & 6.25   & 15.22  & 47.79       \\
C3 & \hspace{1em}+ Cross-CTC      & 6.12   & 14.73   & 46.92   \\
C4 & \hspace{1em}+ SSHR            & \textbf{6.09}   & \textbf{14.05}  & \textbf{45.38}  \\ \bottomrule
\end{tabular}}
\end{table}


\subsection{Analysis}

\subsubsection{Layer-wise analysis}

\begin{figure}[ht]
  \centering
  \includegraphics[width=\linewidth]{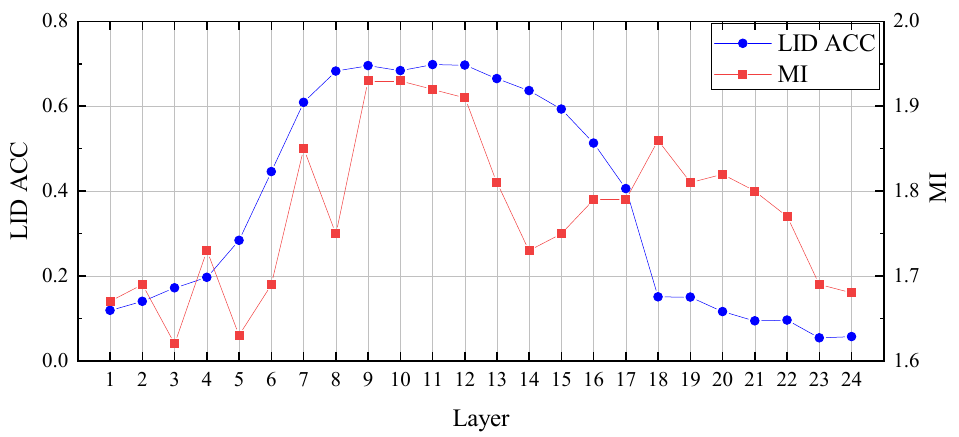}
  \caption{\centering ACC with LID labels and MI with phoneme labels.}
  \label{fig:layer_anaylize}
\end{figure}

Fig.\ref{fig:layer_anaylize} shows the results of the layer-wise analysis of the MMS model.
The x-axis in the figure represents the 24 layers of the MMS. Meanwhile, the y-axis for the blue line indicates the LID prediction accuracy for each MMS layer. We conduct classification experiments using the ML-SUPERB~\cite{23MLSUPERB}, and the findings indicate that MMS exhibits greater language-related information relevance in the middle layers, specifically layers 8 to 12.
Regarding the red line, the y-axis signifies each MMS layer's MI with phoneme labels, which can represent content-related information. We conduct experiments utilizing the Common Voice~\cite{19commonvoice}, revealing that MMS demonstrates heightened content-related information relevance in the middle and high layers but gradually diminishes in the final layers. 

\subsubsection{Extracting Language-related Information}
We explore various LID prediction methods and assess the impact of LID layers on downstream multilingual ASR task. From the Table~\ref{tab:extract-lang-info}, D2 does not use the LID-related frame presented in Section \ref{sc:3,3}, but uses ground truth with LID labels for comparing with D3. D3 is the use of LID-related frames at layer 3, which is an improvement over D2, demonstrating the effectiveness of LID-related frames. D1 implement a strategy akin to that employed in HierLID~\cite{23HierCTC}, utilizing CTC loss for frame-level LID prediction and alignment in layer 3. Compared to D1, the improvement of D3 may be attributed to our deliberate avoidance of errors from LID prediction. Moreover, C1 extracts the LID-related frame at layer 8 of higher language-related information compared to D3, which brings further gains. Further investigation, as presented in C1, highlights that extracting the LID-related frame from more language-related layers decline PER.
\begin{table}[h]
\centering
\caption{PER (\%) results on Common Voice for methods focused on Extracting Language-related Information. CV-PER represents PER on Common Voice.}
\label{tab:extract-lang-info}
\begin{tabular}{@{}lll@{}}
\toprule
ID & Model & CV-PER \\ \midrule
B0 & Baseline & 6.70 \\
D1 & LID$_{\text{tok}}$ in layer 3~\cite{23HierCTC} & 6.63 \\
D2 & No LID-related frame & 6.68 \\
D3 & LID-related frame, layer 3 & 6.51 \\
C1 & LID-related frame, layer 8 & \textbf{6.38} \\ \bottomrule
\end{tabular}
\end{table}
\vspace{-10pt}

\subsubsection{Refining SSL Parameters}
Table~\ref{tab:refine-ssl-params} presents the influence of various initialization methods on the outcomes. Among them, E1 corresponds to the approach proposed in previous work~\cite{21layerwise}, which involves re-initializing the parameters of the last three layers. On the other hand, E2 is the technique outlined in Section~\ref{sc:3,3}, wherein the parameters within the final three layers (layers 22, 23, and 24) are substituted with more informative parameters from the intermediate three layers (layers 19, 20, and 21). The experimental findings highlight that E2 does not perform better on the Common Voice dataset than E1. C2 constitutes another approach introduced in Section~\ref{sc:3,3}, involving the omission of parameters in the last three layers. Compared to E1, C2 performs better while boasting a reduced parameter count. We also show the method of deleting the last four layers in E3, which has a comparable performance. Insights drawn from the results of the Common Voice dataset indicate that, for MMS, the parameters of these final three layers can be omitted.

\begin{table}[h]
\centering
\caption{PER (\%) results on Common Voice for methods focused on Refining SSL Parameters. CV-PER represents PER on Common Voice.}
\label{tab:refine-ssl-params}
\begin{tabular}{@{}lll@{}}
\toprule
ID & Model & CV-PER \\ \midrule
B0 & Baseline & 6.70 \\
E1 & Random init last 3 layers~\cite{21layerwise} & 6.44 \\
E2 & Replace middle 3 layers & 6.52 \\
E3 & Delete last 4 layers & 6.26 \\
C2 & Delete last 3 layers & \textbf{6.25} \\ \bottomrule
\end{tabular}
\end{table}
\vspace{-5pt}


\subsubsection{Enhancing Content-related Information}
Table~\ref{tab:enhance-content-info} demonstrates the application of diverse intermediate layer techniques to enhance content-related information within the posterior n-layers. In this context, F1 embodies the GIC~\cite{22GIC} method in layer 21. F2 corresponds to the Cross-CTC presented in Section \ref{sc:3,4} in layer 21 and has an improvement compared to F1. We further conduct experiments incorporating multiple layers of GIC in layers 15, 18, and 21, denoted as F3. Remarkably, our observations indicate that adopting the multiple layer strategy from GIC led to a decline in performance. This decline might be due to the current layer being more distant from the final layers and the need for more content-related information. Consequently, C3 opt layers 18, 20, and 22  with more content-related information for the Cross-CTC. While C3 shows almost no improvement relative to F2, our results on ML-SUPERB do show a relative improvement of close to 5\%. The Cross-CTC method leads to an approximately 9\% reduction in PER relative to the baseline model. Furthermore, the choice of multiple layers also plays a crucial role.

\begin{table}[h]
\centering
\caption{PER (\%) results on Common Voice for methods focused on Enhancing Content-related Information. CV-PER represents PER on Common Voice.}
\label{tab:enhance-content-info}
\begin{tabular}{@{}lll@{}}
\toprule
ID & Model & CV-PER \\ \midrule
B0 & Baseline & 6.70 \\
F1 & GIC in layer 21~\cite{22GIC} & 6.25 \\
F2 & Cross-CTC in layer 21 & 6.13 \\
F3 & GIC in layer 15 18 21 & 6.57 \\
C3 & Cross-CTC in layer 18 20 22 & \textbf{6.12} \\ \bottomrule
\end{tabular}
\end{table}
\vspace{-5pt}


\section{Conclusion}
\label{sec:conclusion}
This paper explores the potential of leveraging hierarchical representations in the MMS model to improve multilingual ASR tasks. We address the problems of language identification and content enhancement through the LID-related frame and Cross-CTC methods. Our proposed SSHR demonstrates discernible improvements across various experiments, outperforming the baseline and achieving state-of-the-art results on Common Voice and ML-SUPERB datasets. Our findings highlight the importance of utilizing MMS's hierarchical representations for multilingual ASR tasks. The advancements presented in this work provide promising directions for further research in downstream tasks. It is important to note that our experiments have only been validated with low-resource fine-tuning. In the future, we will evaluate the potential of SSHR when applied to high-resource data.

\vfill\pagebreak

\bibliographystyle{IEEEbib}
{\fontsize{9pt}{11pt}\selectfont
\bibliography{ref}
}
\end{document}